\pdfoutput=1
\documentclass[citeauthoryear]{llncs}

\usepackage{geometry}
\usepackage{graphicx}
\usepackage{amsmath,amssymb} 
\usepackage{color}

\usepackage{times}
\usepackage{epsfig}
\usepackage{eufrak}

\DeclareMathOperator{\Binomial}{B}
\DeclareMathOperator{\Pa}{Pa}

\begin{document}


\mainmatter
\pagestyle{plain}

\title{Stability and Structural Properties of Gene Regulation Networks with Coregulation Rules (Preprint)} 


\author{Jonathan H. Warrell\inst{1}\inst{2}$^*$, Musa M. Mhlanga\inst{1}\inst{2}\inst{3}}

\institute{{\em Gene Expression and Biophysics group, Council for Scientific and Industrial Research, Pretoria, South Africa}
\and {\em Division of Chemical Systems and Synthetic Biology, Faculty of Health Sciences, \\ University of Cape Town, South Africa}
\and {\em Unidade de Biofisica e Express\~{a}o Gen\'{e}tica, Instituto de Medicina Molecular, Universidade de Lisboa, Portugal} \\ \quad \\ *jonathan.warrell@gmail.com}

\maketitle

\begin{abstract}
Coregulation of the expression of groups of genes has been extensively demonstrated empirically in bacterial and eukaryotic systems.  Such coregulation can arise through the use of shared regulatory motifs, which allow the coordinated expression of modules (and module groups) of functionally related genes across the genome.   Coregulation can also arise through the physical association of multi-gene complexes through chromosomal looping, which are then transcribed together.  We present a general formalism for modeling coregulation rules in the framework of Random Boolean Networks (RBN), and develop specific models for transcription factor networks with modular structure (including module groups, and multi-input modules (MIM) with autoregulation) and multi-gene complexes (including hierarchical differentiation between multi-gene complex members).  We develop a mean-field approach to analyse the stability of large networks incorporating coregulation, and show that autoregulated MIM and hierarchical gene-complex models can achieve greater stability than networks without coregulation whose rules have matching activation frequency.  We provide further analysis of the stability of small networks of both kinds through simulations.  We also characterize several general properties of the transients and attractors in the hierarchical coregulation model, and show using simulations that the steady-state distribution factorizes hierarchically as a Bayesian network in a Markov Jump Process analogue of the RBN model.
\end{abstract}

\section{Introduction}\label{sec:intro}

Empirical work has demonstrated that coregulation is a ubiquitous characteristic of transcription factor networks (which we shall refer to as gene regulation networks, or GRNs) in bacterial and eukaryotic systems.  The operon model (\cite{jacob_61}) provides a simple example of the coregulation of a group of genes in bacteria.  Extensive investigation of GRNs in the bacteria {\em Escherichia coli} and the yeast {\em Saccharomyces cerevisiae} has revealed a range of network motifs indicative of various types of coregulation between groups of operons or genes (\cite{alon_06,alon_07,milo_02,shenn_orr_02}).  Further investigation of {\em Saccharomyces cerevisiae} has revealed that a large component of its GRN can be modeled as a collection of modules each with characteristic functional roles and regulatory motifs, and that these can be further organized into a collection of {\em module groups} which share common regulators and regulatory motifs in a combinatorial fashion (\cite{peer_02,segal_02,segal_03}).

Further work in eukaryotic cells has revealed that coregulation of gene transcription can also occur through the physical association of multi-gene complexes which are transcribed together (\cite{fanucchi_13,li_12,papantonis_12}).  Such coregulation is dependent on the 3D chromatin conformation of the nucleus in a given cell, which can be stochastic (\cite{fanucchi_13}).  Further, results have shown that the multi-gene complexes formed can have a hierarchical structure, where the transcription of certain members in the complex is dependent on other members in the multi-gene complex being cotranscribed (\cite{fanucchi_13,li_12}).  The functional relevance of such hierarchical coregulation is yet to be characterized.

In the following, we are interested in characterizing the properties of networks involving coregulation rules in a general sense.  For this purpose, we draw on the framework of Random Boolean Networks (RBNs), or {\em NK}-networks (\cite{kauffman_69,kauffman_93}).  RBNs have provided a powerful framework in which the relationship between topological/rule-based constraints and dynamic (or emergent) properties of a network can be characterized through a combination of analytic and simulation-based methods. Examples of biologically inspired constraints include scale-free topology (\cite{aldana_03,kauffman_04}) and {\em canalyzing} regulation rules (\cite{kauffman_04}); while emergent dynamical properties include network stability/criticality (\cite{derrida_86,kauffman_04}), attractor and transient structure (\cite{kauffman_93,kauffman_04}), and evolvability (\cite{torres_12}).  The application of mean-field methods from statistical physics has permitted analytic results to be obtained in a number of these cases (\cite{aldana_03,derrida_86,kauffman_04}).  Further, empirical results have supported the general relevance of such models to the dynamics of actual biological systems, despite the simplification involved in the Boolean assumption (\cite{huang_05,chang_08}).

We introduce a general formulation of coregulation rules in the RBN context, along with a characterization of the mean-field dynamics of models incorporating such rules, which can be used to analyse the dynamics in networks incorporating any of the types of coregulation described above. We focus particularly on (a) the {\em multi-input module} (MIM) network motif, which consists of a group of genes, all of which share the same regulators and regulatory logic, and occurs in bacterial and eukaryotic networks (\cite{shenn_orr_02}), and (b) {\em hierarchical coregulation rules}, which provide a model for the kind of hierarchical dependencies described above, observed to occur in multi-gene complexes in some eukaryotic systems (\cite{fanucchi_13,li_12}).  By using the mean-field approach, we show that for certain forms of each type of rule we can demonstrate that networks incorporating coregulation achieve greater stability than can be achieved in networks without coregulation, whose rules have matching activation frequency, as defined below.  We use simulations to verify the conclusions reached by mean-field analysis for small-scale networks.  Further, we characterize general properties of the transients and attractors in the hierarchical coregulation model and the steady-state distribution of its associated annealed model.  Particularly, the steady-state distribution necessarily factorizes hierarchically as a Bayesian network, and we demonstrate through simulations that this property also holds for the steady-state distributions of a Markov Jump Process analogue of the model with hierarchical coregulation rules (using the approach of \cite{gillespie_07}).  We discuss the functional relevance of these properties below.

Section \ref{sec:rbns} offers a brief review of the RBN framework, followed by our general formulation of coregulation rules and their mean-field dynamics in Section \ref{sec:coreg}.  We then investigate the properties of the two particular coregulation models discussed above; the multi-input module in Section \ref{sec:mim}, and hierarchical coregulation model in Section \ref{sec:hiercoreg}.  Section \ref{sec:disc} concludes with a discussion.

\section{Random Boolean Networks}\label{sec:rbns}

We begin by outlining the Random Boolean Network (or $NK$ network) model introduced in \cite{kauffman_69}, and summarize the mean-field approach used to analyse the stability of this model in \cite{derrida_86}.  An RBN model has two parameters, $N$ the number of nodes in the network, which represent the genes, and $K$ the number of regulators for each gene.  At a given time $t$ (where $t\in\mathbb{N}\cup\{0\}$, a gene may be either on or off (transcribed or not transcribed), represented by the Boolean values 1 and 0 respectively.  The binary variable $\sigma_i(t)$ indicates the state of the $i$'th gene at time $t$.  We will write $R_i(k)$ for the $k$'th regulator of gene $i$, where $R_i:\{1...K\}\rightarrow\{1...N\}$.  The dynamics of the model are then fixed by specifying a separate Boolean update rule for each gene:
\begin{eqnarray}\label{eq:rbn1}
\sigma_i(t+1) = f_i(\sigma_{R_i(1)}(t),\;\sigma_{R_i(2)}(t),...,\sigma_{R_i(K)}(t)),
\end{eqnarray}
where $f_i:\mathbb{B}^K \rightarrow \mathbb{B}$, with $\mathbb{B}=\{0,1\}$ (we note that the model incorporates only one layer of regulation, and hence cannot distinguish between pre- and post-transcriptional regulation; related models have attempted to incorporate further regulatory layers, see \cite{markert_10}). Each $(N,K)$ thus fixes a class of networks, which may be sampled randomly by (a) selecting for each gene $i$ its $K$ regulators by sampling $K$ independently identically distributed values from $1...N$, hence setting $R_i$, and (b) for each gene sampling $2^K$ Boolean values to set the output of $f_i$ on each of the possible settings of its regulators.  The outputs to $f_i$ may be sampled by giving even weight to the values $\{0,1\}$, hence performing $2^K$ Bernoulli trials with mean $1/2$; alternatively, a further parameter $p\in[0\;1]$ is introduced (the {\em activation frequency}), and the outputs are sampled via $2^K$ Bernoulli trials with mean $p$.

To analyse the behaviour of the $NK$-model for the case of large networks (as $N\rightarrow\infty$), \cite{derrida_86} introduce an annealed stochastic approximation, which, instead of fixing the regulator indices $R_i$ and update function $f_i$ for a given gene at all time-steps, re-samples them at each time-step using one of the underlying generating distributions discussed above.  To analyse the stability of a given network class, \cite{derrida_86} consider the change in Hamming distance over time between parallel runs of the same network with different initial conditions, under mean-field dynamics in the annealed model.  The mean-field dynamics can be characterized by introducing random variables $x_i(t)\in[0\;1]$, representing the probability that $\sigma_i(1,t)\neq\sigma_i(2,t)$, where $\sigma_i(j,t)$ is the value of $\sigma_i(t)$ when the network is started from initial condition $j$ ($j\in\{1,2\}$).  This gives rise to the following mean-field updates (in a model with $p$ as above, and
$\delta(i,t)=[\sigma_i(1,t)\neq\sigma_i(2,t)]$ where $[A]$ is the Iverson bracket which is 1 when $A$ is true and 0 otherwise, $R_{it}$ and $f_{it}$ are the regulator indexing and update functions respectively sampled for gene $i$ at time $t$, and $S_i(t)=\sum_{j=1...K}\delta(R_{it}(j),t)$):
\begin{eqnarray}\label{eq:rbn2}
x_i(t+1) &=& \sum_{R_{it},f_{it}}P(R_{it})P(f_{it})(P(S_i(t)>0)P(\delta(i,t+1)=1|S_i(t)>0) + \nonumber \\
&& P(S_i(t)=0)P(\delta(i,t+1)=1|S_i(t)=0)) \nonumber \\
&=& 2p(1-p)\cdot\left(1-\sum_{R_{it}}P(R_{it})\prod_{j=1...K}(1-x_{R_{it}(j)}(t))\right) + 0.
\end{eqnarray}
We assume that the $x_i(0)$'s are initialized identically to $p_0$, so that each gene takes a different value in the initial configurations with a probability $p_0$.  Then, by symmetry $x_i(t=1)=x_j(t=1)$ for all $i,j$, and similarly for $t=2...\infty$.  Hence, we can write the updates in Eq. \ref{eq:rbn2} in terms of a single variable, $x(t)=x_i(t)$ for arbitrary $i$:
\begin{eqnarray}\label{eq:rbn3}
x(t+1) &=& 2p(1-p)(1-(1-x(t))^K),
\end{eqnarray}
where we have used the fact as $N\rightarrow\infty$, the probability that each gene has $K$ distinct regulators tends to 1. The fixed points of Eq. \ref{eq:rbn3} can be characterized by considering the function $g(x)=2p(1-p)(1-(1-x)^K)$.  Clearly, for any fixed point, $x=g(x)$, and hence fixed points occur at intersections of the curves $y=x$ and $y=g(x)$.  $y=x=0$ is always such a solution.  Further, we observe that $y=g(x)$ is concave increasing (since $(1-x(t))^K$ is convex decreasing), and hence a second fixed-point will only occur when $g'(0)>1$ (where $g'=\text{d}g/\text{d}x$), which can be expressed as:
\begin{eqnarray}\label{eq:rbn4}
2Kp(1-p) > 1.
\end{eqnarray}
The fixed point 0 is attracting (and hence the network is stable) if and only if Eq. \ref{eq:rbn4} does not hold, since when $2Kp(1-p) > 1$ there is no neighbourhood of 0 such that $g(x)<x$ for all values. For $p=1/2$, this leads to the critical value $K=2$, as observed in \cite{derrida_86} and \cite{kauffman_93}.

\section{Coregulation Rules}\label{sec:coreg}

We first give a general formulation for modeling coregulation in the RBN framework, and provide general methods for studying the mean-field dynamics of such networks.  As above, $N$ and $K$ denote the number of nodes (genes) and regulators respectively.  We further partition the nodes into $G$ groups, which will represent groups of genes which are coregulated.  We assume for convenience that $G$ divides $N$ exactly, so we have $M=N/G$ nodes per group.  We let $C:\{1...N\}\rightarrow(\{1...G\}\times\{1...M\})$ be an arbitrary 1-1 mapping, which we shall consider fixed throughout (without loss of generality), such that $C(i)=(g,m)$ can be read as `node $i$ is the $m$'th member of group $g$'.  A coregulated $NK$-network is then defined by: (1) for each group $g\in\{1...G\}$ picking a regulator indexing function $R_g:\{1...K\}\rightarrow\{1...N\}$ (hence all members of group $g$ share the same regulators); and (2) for each group $g$ picking a coregulation update rule, $f_g:\mathbb{B}^K\rightarrow\mathbb{B}^M$.  The network dynamics can be expressed as:
\begin{eqnarray}\label{eq:cbn1}
\sigma_i(t+1) = f_{C(i)}(\sigma_{R_{g(i)}(1)}(t),\;\sigma_{R_{g(i)}(2)}(t),...,\sigma_{R_{g(i)}(K)}(t)),
\end{eqnarray}
where we write $f_{(g,m)}$ for the projection of $f_g$ onto its $m$'th output, and $g(i)$ for the projection of $C(i)$ onto its first output.  We will assume that $R_g$ and $f_g$ are sampled independently for each group.  Hence, we will consider distributions over a class of coregulated networks which factor as $P(R,f) = \prod_g(P(R_g)P(f_g))$.

As in Section \ref{sec:rbns}, we introduce an annealed version of the coregulated model in order to study its mean-field dynamics.  We therefore allow $R$ and $f$ to be resampled at each time step from $P(R,f)$ as above, and write $R_{gt}$ and $f_{gt}$ for the regulation index and update function respectively sampled for group $g$ at time $t$.  As above, we consider the situation that the same network is started at two different initial states, and are interested in whether the Hamming distance converges to 0 as an indicator of network stability.  We write $\sigma_i(j,t)$ for the state of node $i$ at time $t$ started at the $j$'th initial condition ($j\in\{1,2\}$), and $\sigma_g(j,t)$ for the vector of states of the nodes in group $g$ at time $t$ starting at the $j$'th initialization.  For the mean-field dynamic analysis, we introduce continuous variables $c_g(v,w,t)$, where $v,w \in \mathbb{B}^M$ are all possible settings of $\sigma_g(1,t)$ and $\sigma_g(2,t)$ respectively, and $c_g(v,w,t)=P([\sigma_g(1,t)=v]\wedge[\sigma_g(2,t)=w])$.  The variable $x(t)$, representing the expected normalized Hamming distance at time $t$ as in Section \ref{sec:rbns}, can be calculated as $x(t)=(1/(GM))\sum_{gvw} c_g(v,w,t)H(v,w)$, where $H(v,w)$ is the Hamming distance between $v$ and $w$.  In the general formulation as given, the mean-field updates can be calculated as:
\begin{eqnarray}\label{eq:cbn2}
c_g(v,w,t+1) &=& \sum_{R_g,f_g} P(R_g)P(f_g)\sum_{v',w'\in\mathbb{B}^K}c_{R_g}(v',w',t)\cdot\nonumber \\
&&([f_{g}(v')=v]\wedge[f_{g}(w')=w]).
\end{eqnarray}
Here, we us the short-hands $\sigma_{R_g}(j,t)=[\sigma_{R_g(1)}(1,t),...,\sigma_{R_g(K)}(1,t)]$, and $c_{R_g}(v',w',t)=P([\sigma_{R_g}(1,t)=v']\wedge[\sigma_{R_g}(2,t)=w'])$; writing $\sigma_{R_g}(j,t)$ for the vector of the states of the regulators of group $g$ at time $t$ for initialization $j$.

By introducing further assumptions about the forms of $P(R)$ and $P(f)$, we can derive special forms of the updates in Eq. \ref{eq:cbn2} which can be applied to the models in Sections \ref{sec:mim} and \ref{sec:hiercoreg}, allowing analytic results to be derived more directly.  We first consider constraining $P(R)$.  We will write $\mathbb{P}(G)$ for the set of permutations on the set $\{1...G\}$, and $\mathbb{P}(G,g,h) = \{\pi\in\mathbb{P}(G)|\pi(g)=h\}$ (the set of permutations mapping $g$ to $h$).  Then, we will say that a distribution $P(R)$ has {\em permutational invariance} iff for any $\pi\in\mathbb{P}(G,g,h)$ and functions $R_g$ and $R_h$, whenever $\forall k [R_g(k)=\pi(R_h(k))]$ (writing $\pi(n)$ for $(\pi(g(n)),m(n))$), we have $P(R_g)=P(R_h)$ (alternatively, the distribution remains invariant up to permutations of the group indices applied jointly to the $R$ subscripts and $R$ output regulator indices).  Further, we call a distribution $P(R)$ {\em simple} when it places zero probability on any function $R_g$ for which $g(R_g(k))=g(R_g(l))$ and $k\neq l$ (all regulators have distinct groups).  We then have:

\vspace{0.3cm}
\noindent\textbf{Proposition 1.} \textit{For a coregulated $NK$-network with a simple distribution $P(R)$ with permutational invariance as above, and assuming $P(f_g)=P(f_h)$ for all $g,h$, and at initialization that $c_g(v,w,0)=c_h(v,w,0)$ for all $g,h$, the mean-field dynamics are such that $c_g(v,w,t)=c_h(v,w,t)=c(v,w,t)$ for all $g,h,t$, where:}
\begin{eqnarray}\label{eq:cbn3}
c(v,w,t+1) &=& \sum_{R',f} P(R')P(f)\sum_{v',w'\in\mathbb{B}^K}(\prod_k c_{R'(k)}(v'_k,w'_k,t))\cdot\nonumber \\
&&([f(v')=v]\wedge[f(w')=w]),
\end{eqnarray}
\noindent\textit{and $R':\{1...K\}\rightarrow\{1...M\}$, $P(R')=\sum_{R_g}P(R_g)[\forall{k}(m(R_g(k))=R'(k))]$ (for arbitrary $g$), $c_{m}(a,b,t)=\sum_{v,w}c(v,w,t)[v_m=a\wedge w_m=b]$.}

\vspace{0.3cm}
\noindent For a proof of Proposition 1, see Appendix A.  We now consider a further restricted class of coregulation networks, in which $P(R)$ is uniform over the set of functions for which $g(R_g(k))\neq g(R_g(l))$ whenever $k\neq l$ (so $P(R)$ remains simple; we shall call a distribution $P(R)$ with this stronger property {\em homogeneous}).  This implies that $P(R')$ (defined in terms of $P(R)$ as in Proposition 1) is uniform.  Further, we assume that $P(f)$ factorizes across its input values: hence, $P(f) = \prod_{v\in\mathbb{B}^K}P'(f(v))$, where $P'(.)$ is independent of $v$. Then, we have:

\vspace{0.3cm}
\noindent\textbf{Proposition 2.} \textit{For a coregulated $NK$-network as in Proposition 1, where in addition $P(R)$ is homogeneous and $P(f)$ factorizes across its input values as $\prod_{v\in\mathbb{B}^K}P'(f(v))$, under mean-field dynamics we have:}
\begin{eqnarray}\label{eq:cbn6}
x(t+1) = \mathcal{K}(1-(1-x(t))^K),
\end{eqnarray}
\noindent\textit{where, writing $H(v,w)$ for the Hamming distance between $v$ and $w$,}
\begin{eqnarray}\label{eq:cbn7}
\mathcal{K} = \frac{1}{M}\sum_{v,w\in\mathbb{B}^m}P'(v)P'(w)H(v,w).
\end{eqnarray}

\vspace{0.3cm}
\noindent \textbf{Proof.}  We assume that we can calculate $x(t)$, and show that the update in Eq. \ref{eq:cbn6} follows.  Following the definition of $x(t)$ above, and using the fact (from Proposition 1) that $c_g(v,w,t+1)=c(v,w,t+1)$ for all $g$:
\begin{eqnarray}\label{eq:cbn8}
x(t+1) = \frac{1}{M}\sum_{v,w\in\mathbb{B}^m}c(v,w,t+1)H(v,w).
\end{eqnarray}
Since $H(v,w)=0$ when $v=w$, these terms will not contribute to the summation in Eq. \ref{eq:cbn8}.  Assuming then that $v\neq w$, following Eq. \ref{eq:cbn3} we have:
\begin{eqnarray}\label{eq:cbn9}
c(v,w,t+1) &=& \sum_{R',f} P(R')P(f)\sum_{v',w'\in\mathbb{B}^K}(\prod_k c_{R'(k)}(v'_k,w'_k,t))\cdot\nonumber \\
&&([f(v')=v]\wedge[f(w')=w])\nonumber \\
&=& \sum_{f,v'\neq w'} x^{K-H(v',w')}(1-x)^{H(v',w')} \cdot \nonumber \\
&& P(f)([f(v')=v]\wedge[f(w')=w])\nonumber \\
&=& \sum_{v'\neq w'} x^{K-H(v',w')}(1-x)^{H(v',w')} P'(v)P'(w)\nonumber \\
&=& (1-(1-x(t))^K) P'(v)P'(w),
\end{eqnarray}
using the fact that only non-matching $v',w'$ values can lead to a non-matching $v,w$ and the homogeneity of $P(R)$ to make the first rearrangement, and the factorization property of $P(f)$ to make the second.  The proposition follows by substituting Eq. \ref{eq:cbn9} into Eq. \ref{eq:cbn8}.

\begin{flushright}
$\square$
\end{flushright}

\section{Multi-Input Module Model}\label{sec:mim}

We first outline two models based on the {\em Multi-Input Module} (MIM) network motif (\cite{shenn_orr_02}).  The first is a generalization of the motif which allows module groups and combinatorial regulation as in \cite{segal_03}, and reduces to the basic motif when only one module per group is allowed.  The second allows a {\em Single-Input Module} (SIM) with autoregulation (\cite{shenn_orr_02}) to be embedded in the MIM motif, as observed in \cite{segal_03}, which reduces to the SIM with autoregulation when $K=1$.

\subsection{Model with Module Groups}\label{sec:mim1}

A multiple-input module is defined as a group of genes whose regulators are identical, and which respond identically to those regulators.  We generalize this motif to a {\em multiple-input module group}, which consists of a group of MIMs each with identical regulators, such that for a given setting of the regulators, either no module is activated, or with probability $p$, each module in the group is activated with a probability $q$. We note that this motif resembles the module group model of \cite{segal_03}, where (1) for simplicity we assume all modules in a module group share exactly the same regulators (rather than all sharing at least one regulator as in \cite{segal_03}), (2) combinatorial use of a common set of regulatory motifs is modelled by activating subsets of the modules (or none) for a given setting of the regulators (the expected size of the subset being determined by $q$), (3) for convenience we assume there is no variation in response within a module, and (4) we allow arbitrary decision rules for the regulatory logic of a given module, in place of decision trees ((3) may be straightforwardly incorporated into our model, by introducing a further parameter representing the probability that a gene is activated given it is in an activated module).

\begin{figure*}[t]
\begin{center}
\includegraphics[width = 0.90\columnwidth]{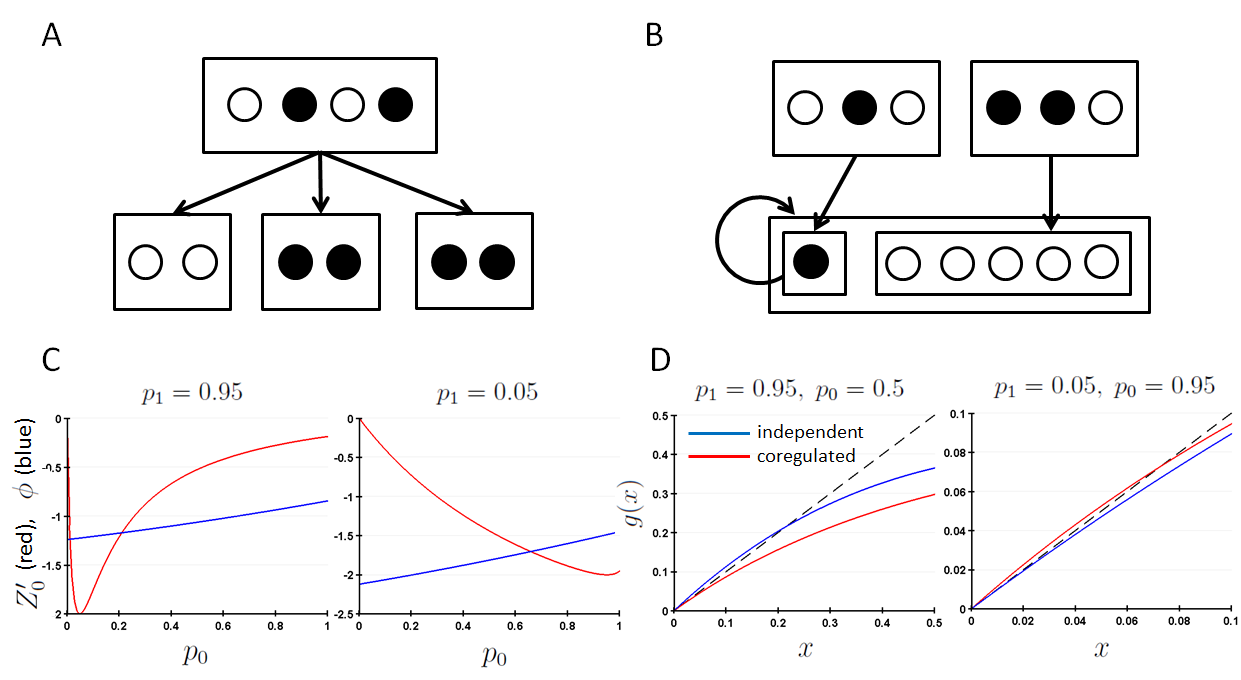}
\end{center}
\caption{{\bf Multi-input module coregulation models.} {\bf (A)}  Model with module groups.  Arrows denote regulatory relationships, and arrows between boxes are implicitly repeated for every pair of nodes in the source and target box.  Illustration shows model with $K=4$ regulators, $L=3$ modules per group, and $N_L=2$ genes per module.  Black/white nodes represent on/off gene states respectively, and a possible network configuration is shown.  {\bf (B)}  Model with autoregulation.  Notation as for (A).  A model is shown with parameters $K=4,L=1,N_L=5,M=6$.  {\bf (C)}  Graphs plot LHS and RHS quantities from Eq. \ref{eq:mim8}.  Parameter ranges where $Z'(0)>\phi$ are those in which the autoregulated MIM model achieves potentially greater stability than an independently regulated model with matching activation frequency, and in ranges where $Z'(0)<\phi$ the independent model achieves potentially greater stability.  Parameters as shown, with $K=3,M=5$, and $p=0.7$ (left), and $p=0.85$ (right).  {\bf (D)}  Graphs show parameter settings for which the autoregulated MIM model model achieves strictly greater stability than the independent model (left), and for which the independent model achieves strictly greater stability (right).  Dotted line shows $y=x$, and an intersection of $g(x)$ with $y=x$ at a non-zero value indicates an attracting fixed point which prevents $x$ reaching $0$ ($g(x)$ is as in Eq. \ref{eq:appB3} for autoregulated model, and as defined following Eq. \ref{eq:rbn3} for the independent model).  Remaining parameters are as for (C).
}
\label{fig1}
\end{figure*}

The multiple-input module group as above may be modelled as a coregulated $NK$-network, as in Section \ref{sec:coreg}.  Here, we take the module-group as the coregulation unit, so that $G$ represents the number of module groups, $M$ the number of genes per module group, and in addition we introduce $L$, the number of modules per module group.  Each module therefore has $N_L=M/L$ genes (which cannot be shared between modules, as in \cite{segal_03}).  We assume that $P(R)$ is simple homogeneous, as defined in Section \ref{sec:coreg}.  With $p$ and $q$ as above, we can define $P(f)$ as:
\begin{eqnarray}\label{eq:mim1}
P(f) &=& \prod_{v\in\mathbb{B}^K} P'(f(v)) \nonumber\\
P'(u) &=& \begin{cases} (1-p)[|u|_1=0] \;\; + & p\Binomial(|u|_1(L/M);L,q) \\
&\mbox{if } \forall (a,b) \;s.\;t.\; l(a)=l(b), u_a=u_b \\
0 & \mbox{otherwise}. \end{cases}
\end{eqnarray}
where $\Binomial(.;N,p)$ denotes the Binomial distribution with $N$ trials and mean $Np$, $|u|_1=\sum_i u_i$ (the $L_1$ norm of $u$), and we have introduced a fixed function $l:\{1...M\}\rightarrow\{1...L\}$, which assigns module indices to genes within a module group ($(\sum_m [l(m)=a]) = M/L$ for any $a$).  The model is illustrated in schematic form in Figure \ref{fig1}A.  We note that, when $q=1$ or $L=1$, the model reduces to an MIM, where all genes are coactivated with a probability $p$ for a given regulator setting.

Since $f$ factorizes over its input values in Eq. \ref{eq:mim1}, Proposition 2 may be applied to analyse the stability of the model.  Hence, following Eq. \ref{eq:cbn7}:
\begin{eqnarray}\label{eq:mim2}
\mathcal{K} &=& \frac{1}{M}\sum_{v,w\in\mathbb{B}^m}P'(v)P'(w)H(v,w) \nonumber \\
&=& \frac{1}{M}(2p(1-p)Mq + p^2M\cdot2q(1-q) \nonumber \\
&=& 2pq(1-pq),
\end{eqnarray}
where line 2 follows by considering that when pairs $(v,w)$ are sampled from $P'(v)P'(w)$, with probability $2p(1-p)$ one is active and the other not, so that the genes in each module take different values in $v$ and $w$ with probability $q$, while with probability $p^2$ both are active, and so each module takes a different value with probability $2q(1-q)$.  We are interested in comparing the stability of this model with an $NK$ network without coregulation (which we will refer to as an {\em independent} or {\em independently regulated} model), whose outputs are active with the same probability (which we refer to by saying it has the same {\em activation frequency}).  Considering Eq. \ref{eq:mim1}, the expected proportion of active outputs across all inputs to $f$ is $pq$.  Hence, following Eq. \ref{eq:rbn3}, the update for the normalized Hamming distance in an $NK$ net whose outputs are sampled independently with this probability is:
\begin{eqnarray}\label{eq:mim3}
x(t+1) = 2pq(1-pq)(1-(1-x(t))^K),
\end{eqnarray}
which is identical to Eq. \ref{eq:cbn6} with $\mathcal{K}$ as above.  In the limit of large $N$ then, the stability of the multiple-input module group model is not different to a model without coregulation with matching activation frequency.  While in this case coregulation does not affect stability therefore in the asymptotic limit, for small networks we can show that the stability of the coregulated model is greater than the corresponding model with independent regulation (see Appendix B).

\subsection{Model with Autoregulation}\label{sec:mim2}

We now adapt the multi-input module group model above to include autoregulation.  For this purpose, we add a further node to the coregulation unit, so that $M=LN_L+1$, and $l:\{1...M\}\rightarrow\{0...L\}$, with $(\sum_m (l(m)=a))=L_N$ for $a=1...M$, and $l(1)=0$; hence $m=1$ picks out a distinguished member of each module group, which will be subject to autoregulation.  Further, we expand the number of regulators to $2K-1$, which will allow us to select different regulators for the distinguished member versus the other members of the module group.  The distribution of the regulation function is:
\begin{eqnarray}\label{eq:mim4}
P(R_g) &\propto& \begin{cases} 0 &\mbox{if } \exists (a,b),a\neq b, g(R_g(a))=g(R_g(b)) \\
0 &\mbox{if } R_g(1)\neq C(g,1) \\
1 & \mbox{otherwise}. \end{cases}
\end{eqnarray}
The first case of Eq. \ref{eq:mim4} ensures that $P(R_g)$ is simple, while the second ensures that the first regulator of group $g$ is the distinguished member of group $g$ itself (hence, it is not simple homogeneous).  Clearly, Eq. \ref{eq:mim4} has the permutational invariance property discussed previously, since any permutation on $\{1...G\}$ which maps group $g$ to $h$ will map distinguished member $C(g,1)$ to $C(h,1)$ also.  We complete the model definition by specifying $P(f)$.  Here, we use four parameters: $p_0$ is the probability the distinguished member will be activated at $t+1$ given it was inactive at $t$; $p_1$ is the probability it will be active at $t+1$ given it is active at $t$; $p$ is the probability that the module group as a whole will be activated; and $q$ is as above the probability a module in the group is active if the group is activated.  Further, we require that, writing $u_a$ for $u\in\mathbb{B}^{2K-1}$ restricted to indices $[1...K]$ and $u_b$ for $u$ restricted to indices $[1,K+1...2K-1]$: $u_a=v_a$ implies $f_1(u)=f_1(v)$, and $u_b=v_b$ implies $f_{2...M}(u)=f_{2...M}(v)$ (so that for all other functions, $P(f)=0$).  The above leads us to define $P(f)$ as:
\begin{eqnarray}\label{eq:mim5}
P(f) &=& P_a(f')P_b(f''),
\end{eqnarray}
where $f':\mathbb{B}^K\rightarrow \mathbb{B}$, $f'(u_a)=f_1(u)$; $f'':\mathbb{B}^K\rightarrow \mathbb{B}^{M-1}$, $f''(u_b)=f_{2...M}(u)$; and:
\begin{eqnarray}\label{eq:mim6}
P_a(f') &=& \prod_{u_a\in\mathbb{B}^K} P_a'(u_a(1),f'(u_a)) \nonumber\\
P_a'(\beta,v) &=& \begin{cases} (1-p_0)^{[v=0]}p_0^{[v=1]} &\mbox{if } \beta=0 \\
(1-p_1)^{[v=0]}p_1^{[v=1]} & \mbox{otherwise}, \end{cases} \nonumber \\
P_b(f'') &=& \prod_{u_b\in\mathbb{B}^K} P_b'(f'(u_b)) \nonumber\\
P_b'(v) &=& \begin{cases} (1-p)[|v|_1=0] \;\; + & p\Binomial(|v|_1(L/M);L,q) \\
&\mbox{if } \forall (a,b) \;s.\;t.\; l(a)=l(b)>0, v_a=v_b \\
0 & \mbox{otherwise}. \end{cases}
\end{eqnarray}

A specific case of a module group with a regulator subject to autoregulation is observed in \cite{segal_03} (the autoregulator being the transcription factor, Yap6; see their Figures 5 and 7f).  Further, for $L=K=1$, the model reduces to an autoregulated single-input module, which is observed to be a common network motif in bacteria and eukaryotes (\cite{alon_07}).  The model is illustrated in Figure \ref{fig1}B.  For convenience, in Figure \ref{fig1}B and the following, we will take $L=1$, so that each module group contains only one module, and each coregulation group contains a single module of $M-1$ nodes along with one distinguished autoregulatory node (hence forming a coregulation group of size $M$).  We also set $q=1$, so that the module activation is determined entirely by $p$.

As remarked, the distribution $P(R_g)$ as defined in Eq. \ref{eq:mim4} is simple and has the permutational invariance property.  Hence Proposition 1 can be used to analyse the stability of the model.  Further, writing $p'$ for the probability that an arbitrary output of $f$ is 1 under $P(f)$ (the activation frequency), we have:
\begin{eqnarray}\label{eq:mim7}
p' &=& \sum_{f} P(f) (1/M) \sum_{u,m} [f_m(u)=1] \nonumber\\
&=& \sum_{f} P_a(f')P_b(f'')(1/M)\sum_{u,m} [f_m(u)=1] \nonumber\\
&=& \frac{0.5(p_0+p_1)+(M-1)p}{M}.
\end{eqnarray}
Hence, we are interested in the circumstances in which a coregulated model as above is able to achieve (or unable to achieve) greater stability than an independently regulated model with activation frequency $p'$. Using Proposition 1, we can show that:

\vspace{0.3cm}
\noindent\textbf{Proposition 3.} \textit{For an autoregulated MIM model as above, with $L=q=1$ and identical initialization for all coregulated groups, the autoregulated MIM model has greater or equal stability to an independent $NK$-network with identical activation frequency (meaning that $x$ has a fixed point greater than 0 only when the independent model does) when the following condition holds:}
\begin{eqnarray}\label{eq:mim8}
Z'(0) &\geq& \phi(p_0,p_1,p) \nonumber \\
&=& \frac{M\left(2\left(\frac{M-1}{M}\right)(K-1)p(1-p)-2Kp'(1-p')\right)}{1+2(M-1)p(1-p)},
\end{eqnarray}
\noindent \textit{where}
\begin{eqnarray}\label{eq:mim9}
Z'(0) &=& \frac{((1-p_0)p_0-(1-p_1)p_1)(p_1p_0-(1-p_1)(1-p_0))}{(1-(1-p_0)p_1)^2-((1-p_1)p_0)^2}+ \nonumber \\
&& \frac{((1-p_0)p_0+(1-p_1)p_1)((1-p_0)p_0-(1-p_1)p_1-1)}
{(1-(1-p_0)p_1)^2-((1-p_1)p_0)^2},
\end{eqnarray}
\noindent \textit{subject to the condition that $Z(x)$, defined in Appendix C, is decreasing convex over the interval $[0\;1]$.  Similarly, the independent model has greater or equal stability when $Z'(0) \leq \phi$.}

\vspace{0.3cm}
\noindent For a proof of Proposition 3, see Appendix C.  We can use the condition in Eq. \ref{eq:mim8} to investigate the stability of the autoregulated model under various parameter settings.  Figure \ref{fig1}C for instance plots $Z'(0)$ against $\phi(p_0,p_1,p)$ as a function of $p_0$, while fixing $p_1$ to a high (0.95) and low (0.05) value on the left and right graphs respectively ($p$ fixed in both).  In regions in which $\phi>Z'(0)$, the coregulated model cannot be less stable than the independent model, while for $\phi<Z'(0)$ it cannot be more stable.  For $p_1=0.95$, corresponding to positive-feedback, the coregulated model can only achieve greater stability when $p_0$ is not too low: for low values of $p_0$ ($<\sim0.2$), the $0$ value of the autoregulator is a competing stable value, which disrupts the stability of the value $1$ (left).  For $p_1=0.05$, corresponding to negative feedback, the opposite is the case, and the model can achieve greater stability only when $p_0$ reinforces the stability of $0$ (right). Interestingly, in the positive feedback case (left), the coregulation model can be stabilized also for very low values of $p_0$ ($<\sim0.01$), corresponding to the $0$ value of the autoregulator achieving greater stability than the value $1$.  In all cases, the precise points at which the relative changes in stability and instability occur depend on the value of $p$, which determines the stochasticity of the module.  Figure \ref{fig1}D picks parameter settings demonstrating a strict increase and decrease in stability of the coregulated model with respect to the independent model, using particular settings from the corresponding regions in Figure \ref{fig1}C.  We further verify the stability/instability of these models in the small network setting by comparing simulations with the mean-field analysis given above (see Appendix B).

\section{Hierarchical Coregulation Model}\label{sec:hiercoreg}

The formation of multi-gene complexes via chromatin conformation, whose members are cotranscribed, has been ubiquitously observed in eukaryotic nuclei (\cite{papantonis_12}).  Two studies have observed that the members of such complexes may exhibit a hierarchical organization, such that members lower in the hierarchy are only transcribed if members at a higher level are transcribed.  For instance, \cite{li_12} observe a four member complex, in which one member ({\it GREB1}) must be transcribed for transcription of the remaining members to occur. \cite{fanucchi_13} observe a gene-complex with three members, arranged into a three-tier hierarchy such that high-level transcription of the second ({\it TNFAIP2}) depends on transcription of the first ({\it SAMD4A}), and transcription of the third ({\it SLC6AS}) likewise depends on the second.

We present here a simple model of such hierarchical coregulation.  Our model is a coregulated network as above, where we take the coregulation units to represent multi-gene complexes (hence $M$ denotes the number of genes per complex).  For generality, we allow an arbitrary hierarchical structure to be placed over the members of a complex, which is represented by the function $\Pa : \{1...M\}\rightarrow\{0...M\}$.  $\Pa(m_1) = m_2$ denotes that $m_2$ is the parent of $m_1$ in the hierarchical ordering, and we set $\Pa(m) = 0$ for members which have no parent.  We assume that for no element $m_1$ do we have $\Pa(m_1)=m_2,\Pa(m_2)=m_3,...,\Pa(m_n)=m_1$ (there are no loops), and hence $\Pa$ induces a partial ordering over $\{1...M\}$.  For convenience, the function $\Pa$ is fixed for a given network, implying that all coregulation groups share the same hierarchical structure.

\begin{figure*}[t]
\begin{center}
\includegraphics[width = 0.9\columnwidth]{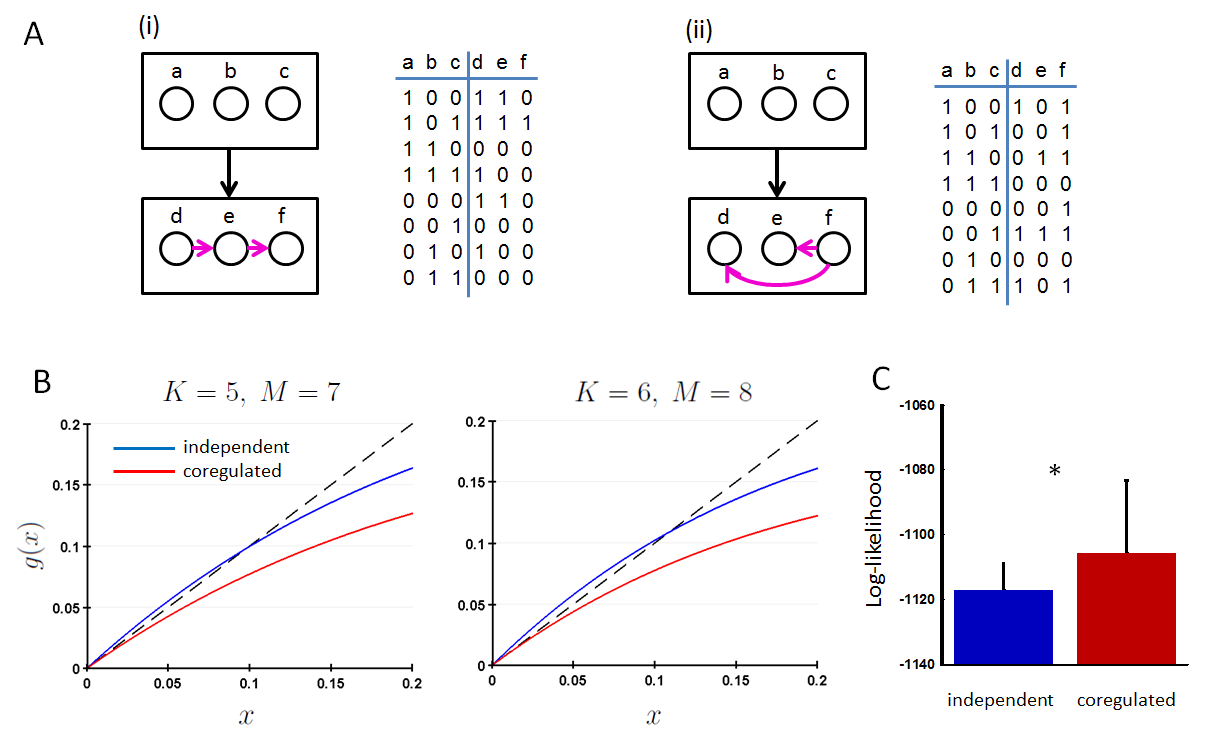}
\end{center}
\caption{{\bf Hierarchical coregulation models.} {\bf (A)} Illustrates possible models.  Notation is as in Fig \ref{fig1}, and additionally magenta arrows indicate $\Pa$ relationship over rule outputs (where the source node of a magenta arrow is the parent of the target node).  For both (i) and (ii), $K=M=3$.  (i) shows a model which places a total order over its outputs ($\Pa(m)=m-1$), while (ii) shows a partial ordering.  Tables show a valid coregulatory function $f$ which respects the hierarchical structure of each model. {\bf (B)} Shows two parameter settings for which the hierarchical coregulation model achieves strictly greater stability than an independent model with matching activation frequency.  A model is stable when $g(x)$ remains below the dotted line $y=x$, preventing a non-zero attracting fixed point from occurring (see Eq. \ref{eq:hcrg8}).  For both graphs, $p=0.5$ and $\Pa(m)=m-1$. {\bf (C)} Compares the fit of a Bayesian network to the steady-state distributions of independent and hierarchically coregulated models, using a Markov Jump process analogue to the Boolean model.  For both models, $N=20,K=2,p=0.5$, and for the coregulated model $M=4$.  The median log-likelihood of the best fitting Bayesian Network is shown for 40 trials, and error bars indicate the upper quartiles (*$p<0.05$, one-tailed Mann-Whitney test).
}
\label{fig2}
\end{figure*}

The model has one additional parameter $p$, which is the probability that a gene is active given its parent is active, or its probability of activation if it has no parent.  For a gene whose parent is inactive, we take its probability of activation to be $0$.  Hence, we specify $P(f)$ as follows:
\begin{eqnarray}\label{eq:hcrg1}
P(f) &=& \prod_{v\in\mathbb{B}^K} P'(f(v)) \nonumber\\
P'(u) &=& \prod_m p'(u(m),u,\Pa(m)) \nonumber \\
p'(b,u,m) &=& \begin{cases} p^{b}(1-p)^{(1-b)}
& \mbox{if } (m\neq 0,\;u(m)=1)\vee(u(m)=0) \\
[b=0] & \mbox{otherwise}. \end{cases}
\end{eqnarray}
$P(f)$ therefore factorizes across its input values, according to our earlier definition (Section \ref{sec:coreg}).  The model is completed by specifying $P(R)$, which we take to be simple homogeneous.  Examples of coregulation rules with various hierarchial structures across their outputs, and possible $f$ functions sampled from the corresponding $P(f)$ distributions are shown in Figure \ref{fig2}A.

\subsection{Stability}

Since $P(f)$ factorizes across its inputs and $P(R)$ is simple homogeneous, mean-field analysis of the stability of a given model with hierarchical coregulation can be made using Proposition 2.  For simplicity, we give a detailed analysis here of the class of hierarchical models which place a total order across their members of the form $\Pa(m) = m-1$, so that the first member of each coregulation group has no parent, the second member is the child of the first, and so on.  For the case of $M=3$, this corresponds directly to the observed hierarchy in \cite{fanucchi_13}.  These rules can also be seen to impose an analogous structure over their outputs to the structure imposed by {\em canalyzing rules} over their inputs (see \cite{kauffman_04}). The relationship between rules of this form and canalyzing rules is discussed further in Appendix D. For this class of models, we have:

\vspace{0.3cm}
\noindent\textbf{Proposition 4.} \textit{For a hierarchical coregulated $NK$-network with $\Pa(m) = m-1$ as above, under mean-field dynamics we have:}
\begin{eqnarray}\label{eq:hcrg2}
x(t+1) = \mathcal{K}(M)(1-(1-x(t))^K),
\end{eqnarray}
\noindent\textit{where,}
\begin{eqnarray}\label{eq:hcrg3}
\mathcal{K}(M) = \frac{2}{M}\sum_{i=1...M}i\left(\sum_{j=0...(M-i-1)}p^{2j+i}(1-p)^2+p^{2M-i}(1-p)\right).
\end{eqnarray}

\vspace{0.3cm}
\noindent \textbf{Proof.}  From Eq. \ref{eq:hcrg1}, we observe that the only valid output values for $f$ are $v_0=[0,0,...,0]$, $v_1=[1,0,...,0]$, $v_2=[1,1,0,...,0]$, ..., $v_M=[1,1,...,1]$.  Further, we know that the Hamming distance between any two outputs is in the set $\{0...M\}$, and that for a given Hamming distance $i$, it can only occur between two outputs of the form $v_j$ and $v_{j+i}$.  Assuming $j+i<M$, the probability of generating $v_j$ (from $P'(.)$ in Eq. \ref{eq:hcrg1}) is $p^j(1-p)$, and the probability of generating $v_i$ is $p^{j+i}(1-p)$.  For the case that $j+i=M$ the probability of generating $v_i$ is $p^{j+i}$.  Hence, for outputs $u,v$ sampled independently from $P'(.)$, we have that:
\begin{eqnarray}\label{eq:hcrg5}
P(H(u,v)=i) &=& \sum_{j=0...M-i-1}2p^{j}(1-p)p^{j+1}(1-p)+2p^{M-i}(1-p)p^M \nonumber\\
&=& 2\sum_{j=0...M-i-1}p^{2j+1}(1-p)^2+p^{2M-i}(1-p),
\end{eqnarray}
where the factor of 2 arises from the fact that each pair $v_j$ and $v_{j+i}$ may be ordered arbitrarily for $H(u,v)>0$.  Since, following Eq. \ref{eq:cbn7} of Proposition 2,
\begin{eqnarray}\label{eq:hcrg6}
\mathcal{K}(M)=\frac{1}{M}\mathbb{E}(H(u,v))=\frac{1}{M}\sum_{i=1...M}i P(H(u,v)=i),
\end{eqnarray}
the proposition follows.

\begin{flushright}
$\square$
\end{flushright}

Following Section \ref{sec:coreg}, we observe that $(1-(1-x(t))^K)$ is concave, and so a model with hierarchical coregulation where $\Pa(m) = m-1$ will have a fixed point only at zero when $g'(0)<1$, where $g'$ is the differential with respect to $x$ of the function $g(x)=\mathcal{K}(M)(1-(1-x(t))^K)$; hence the model is stable when $\mathcal{K}(M)K<1$.  We can calculate the activation frequency (the probability of an arbitrary output gene being on) of the hierarchical model as:
\begin{eqnarray}\label{eq:hcrg7}
p' &=& \frac{1}{M}\sum_{i=1...M}P'(u_i=1) \nonumber\\
&=& \frac{1}{M}\sum_{i=1...M}\prod_{j=1...i}p \nonumber\\
&=& \frac{1}{M}\sum_{i=1...M}p^i
\end{eqnarray}
Hence, by Eq. \ref{eq:rbn4}, a independent model with the same activation frequency is stable when $2(K/M^2)p'(1-p')=2(K/M^2)(\sum_ip^i)(1-\sum_ip^i)<1$.  The hierarchical coregulation model therefore can achieve greater stability than an independent model with identical activation frequency when:
\begin{eqnarray}\label{eq:hcrg8}
\mathcal{K}(M)<2\frac{1}{M^2}(\sum_{i=1...M}p^i)(M-\sum_{i=1...M}p^i)
\end{eqnarray}

For simplicity, we now analyse the case in which $p=0.5$.  Here:
\begin{eqnarray}\label{eq:hcrg9}
\mathcal{K}(M)&=& \frac{2}{M}\sum_{i=1...M}i\left(0.5^{2+i}\sum_{j=0...(M-i-1)}0.5^{2j}+0.5^{2M-i+1}\right)\nonumber\\
&=& \frac{2}{M}\sum_{i=1...M}i\left(0.5^{2+i}\frac{1-0.5^{2(M-i)}}{1-0.5^2}+0.5^{2M-i+1}\right)\nonumber\\
&=& \frac{2\sum_{i=1...M} i0.5^i}{3M} + \frac{0.5^{2M}\sum_{i=1...M} i0.5^{-i}}{3M}\nonumber\\
&=& \frac{2(2-0.5^M(2+M)}{3M} + \frac{0.5^{2M}(2+2^M(2M-2))}{3M},
\end{eqnarray}
using the geometric series expansion $\sum_{j=0...N}p^j=(1-p^N)/(1-p^2)$ in line 2, and the identities $\sum_{k=1...N}k0.5^k=2-0.5^N(2+N)$ and $\sum_{k=1...N}k2^k=2+2^N(2N-2)$ in line 4.  Further, since $\sum_{i=1...N}0.5^i=1-0.5^N$, we have that $p'=(1/M)(1-0.5^M)$, and $(2/M^2)p'(M-p')=(2/M^2)(M-1+0.5^M-0.5^{2M})$.  For the case of $p=0.5$ then, multiplying Eq. \ref{eq:hcrg8} through by $M>0$ and rearranging, the condition for coregulation to have a stabilizing effect becomes:
\begin{eqnarray}\label{eq:hcrg10}
\left(2-\frac{2}{M}-\frac{4}{3}\right) + \left(0.5^{M-1}-\frac{0.5^{2M-1}}{M}-\frac{2(0.5)^{2M}}{3M}\right)+\frac{2(0.5)^M}{M}>0.
\end{eqnarray}
The left-hand side of Eq. \ref{eq:hcrg10} is 0 for $M=1$, and is satisfied for $M=2,3$.  Clearly, when $M>3$ the first group of terms is positive.  Further, for $M>1$, we have:
\begin{eqnarray}\label{eq:hcrg11}
\frac{0.5^{2M-1}}{M}+\frac{2(0.5)^{2M}}{3M} <  \frac{0.5^{2M-1}}{M}+\frac{0.5^{2M-1}}{M}
= \frac{0.5^{2(M-1)}}{M}
< 0.5^{M-1},
\end{eqnarray}
and hence the second group of terms in Eq. \ref{eq:hcrg10} is positive for $M>3$.  It follows that the condition is satisfied for all $M>1$, and hence the hierarchical coregulation model with $p=0.5$ is always at least as stable as a model without coregulation and equal activation frequency.  Examples are shown in Figure \ref{fig2} for two settings of the parameters $K$ and $M$ (with $p=0.5$) under which the coregulation mean-field dynamics tend to 0 while the independent model with equal activation frequency tends to a non-zero fixed-point (implying strictly greater stability for the coregulation model).  Although the above analysis applies only to models with $p=0.5$, we verified by calculation that Eq. \ref{eq:hcrg8} holds for all combinations of parameters $(p,M)$ where $p\in\{0.01, 0.02, ... ,0.99\}$ and $2\leq M\leq10000$, suggesting that this is a general phenomenon.  Further, we compare the mean-field analysis above with small network simulations of the model in Appendix A.

\subsection{Other Structural Properties}

We briefly consider here some additional structural properties of hierarchically coregulated networks.  In Section \ref{sec:attTrans}, we consider the case $p=0.5$ and $\Pa(m)=m-1$ for convenience.

\subsubsection{Attractor and transient structure}\label{sec:attTrans}

We note first that, because of the form of $P(f)$ the dynamics of the hierarchically coregulated model are highly constrained.  To investigate these constraints, we introduce the notion of an {\em output space}, $O$, which is the set of network configurations that result by applying the network's update rules to all possible input configurations.  For the hierarchical model, we have:
\begin{eqnarray}\label{eq:hcrg12}
O = \{[v_{\mathbf{m}(1)},v_{\mathbf{m}(2)},...,v_{\mathbf{m}(G)}]|\mathbf{m}\in\{0...M\}^G\},
\end{eqnarray}
where the $v_m$ are defined as in the proof of Proposition 4.  Eq. \ref{eq:hcrg12} follows since each coregulation group must take a valid output value following an update, so the network state as a whole must be a combination of these.  Hence, we have $|O|=(M+1)^G$.  In contrast, for an $NK$ network without coregulation such that $p\not\in\{0,1\}$, we have that $O=\mathbb{B}^N$, $|O|=2^N=2^{MG}$.

An attractor is either a network state which updates to itself, or a closed cycle of network states under the update rules.  Any attractor must therefore be a subset of the output space, since each state in the attractor must be mapped to either from itself, or another attractor state.  By contrast, a transient, which is a sequence of network states respecting the update rules whose final member belongs to an attractor, may have an initial state which is not in the output space, but all subsequent states must belong to $O$.

The considerations above suggest that part of the increased stability in the hierarchical coregulated model may be a result of the much smaller output space (in contrast to an independent model with matching activation frequency).  As a further consequence, we might expect that the attractors in the hierarchical coregulated model would be shorter in length, since they are constrained to lie in a smaller output space (alternatively, we might expect that the probability of a path in the output space intersecting itself and forming an attractor after a given number of steps is greater in the coregulated model, since the path is constrained to lie in a smaller subspace, although clearly this probability is not determined by $|O|$ alone).  To test this prediction, we ran simulations of hierarchically coregulated networks while varying $M$, the size of the coregulation groups, from $2...5$, and compared the length of the attractors encountered from a uniformly distributed initial state with those encountered in an $NK$ network without coregulation with update rules having matching activation frequency (fixing $N=40$ and $K=3$ in all simulations).  The results are in Table \ref{table1}, which shows a pronounced reduction in attractor lengths in the coregulated model for all parameter settings.

\begin{table}[!ht]
\centering
\caption{{\bf Comparing attractor lengths in hierarchical coregulation and independent models.}  The median and median absolute deviation of the length of the first attractor encountered over 100 simulations is shown.  $p$-values are shown for the one-tailed Mann-Whitney test.  For all simulations, $N=40,K=3$.}
\begin{tabular}{|p{2.5cm}|p{2.5cm}|p{2.5cm}|p{2.5cm}|p{2.5cm}|}
\hline
 & $M=2$ & $M=3$ & $M=4$ & $M=5$ \\ \hline
 Coregulated & $9.6\pm7.6$ & $3.9\pm2.1$ & $2.2\pm1.2$ & $1.8\pm0.8$ \\ \hline
 Independent & $39.2\pm33.2$ & $10.3\pm7.7$ & $6.2\pm4.2$ & $3.9\pm2.5$ \\ \hline
 $p$-value & $2e-7$*** & $1e-3$** & $2e-7$*** & $2e-7$*** \\ \hline
\end{tabular}
\label{table1}
\end{table}

\subsubsection{Steady-state distributions}\label{sec:mjp}

The steady-state of the annealed model with hierarchical coregulation (which involves sampling a new update rule from $P(f)$ for each group at every time step) can be simply expressed:
\begin{eqnarray}\label{eq:hcrg13}
P(\mathbf{x}) = \prod_{g=1...G} \prod_{m=1...M} P''(\mathbf{x}_g(m)|\mathbf{x}_g(\Pa(m))),
\end{eqnarray}
where we write $\mathbf{x}_g(m)$ for the setting of the $m$'th gene in coregulation group $g$ (for convenience we fix $\mathbf{x}_g(0)=1$), and we have $P''(x|1)=p^{x}(1-p)^{1-x}$ and $P''(x|0)=[x=0]$.  Given the partial order constraints on $\Pa$, Eq. \ref{eq:hcrg13} expresses a factorization of the steady-state distribution in the form of a Bayesian network.

We were interested in whether such hierarchical steady-state distributions would similarly emerge in an analogue to the coregulation model above which drops the Boolean constraint on the gene expression levels.  We therefore investigated a {\em Markov Jump process} (MJP) analogue of the hierarchical coregulation model above.  Here, in place of the Boolean state vector, we let $\mathbf{x}\in(\mathbb{N}\cup\{0\})^N$, which can be taken to represent for instance the transcript count associated with each gene at a given point in time (see \cite{wilkinson_09,wilkinson_11}).  The model is parameterized identically to the Boolean model, with parameters $N,K,M,p$ and associated distributions $P(R),P(f)$.  However, rather than using the function $f_g$ to directly update group $g$ at discrete time-steps, updates take place stochastically in continuous time according to rate equations derived from $f$ and $R$.  An MJP analogue to the independent $NK$ network can be derived similarly, and full details of the derivation are given in Appendix E.

To test for hierarchical structure in the steady-state distributions of the MJP analogue model, we generated 40 hierarchically coregulated and independent networks fixing $N=20$, $K=2$, $p=0.5$ in both models and setting $M=4$ in the coregulated model, and simulated each model 20 times for a duration $t=[0\;100]$ (starting from a 0 initial state).  We took the final states of each simulation to represent samples from the steady state distribution (we observed qualitatively that by $t=100$ the behaviour of each variable had typically stabilized), and fitted a Bayesian network to each steady state, using a Markov chain Monte-Carlo approach to learn the structure of the network (\cite{murphy_01}).  We compared the log-likelihoods of the best fitting Bayesian networks for the hierarchically coregulated and independent conditions.  Figure \ref{fig2}C shows the fit to be significantly higher for the coregulated networks, suggesting that such `structured stochasticity' in the steady-state distributions may be a general feature of this class of networks.

\section{Discussion}\label{sec:disc}

We have provided here a general analysis of coregulation in the context of Random Boolean Networks, particularly with respect to the emergent dynamical properties of networks which embody various kinds of coregulatory motifs.  We have provided general tools for the mean-field analysis of stability in networks incorporating coregulatory rules, and applied these tools to two cases of biological interest.  Significantly, we have shown that multi-input module motifs with positive or negative feedback (autoregulation) can enhance network stability within certain parameter ranges, but can have a destabilizing effect in other parameter ranges.  By contrast, our analysis suggests that hierarchical coregulation is stabilizing in all cases.  We also investigate further structural properties of networks with hierarchical coregulation, showing them to have smaller output spaces and shorter attractors than comparable networks without coregulation, and hierarchically structured steady-state distributions in both Boolean and analogous Markov Jump process models.

The findings we present suggest that part of the function of these motifs may consist in the effects they have on global network dynamics.  For instance, while previous analysis of the SIM with autoregulation suggested that feedback can function as a stabilizing mechanism when viewed in isolation (for instance, allowing the system to respond only to signals which are temporally extended, see \cite{alon_06}), the analysis presented here allows us to investigate the effects on global stability of such feedback.  Indeed, our analysis suggests that global context needs to be taken into account in assessing the effects of such motifs.  Similarly, it has been suggested that hierarchical coregulation of multi-gene complexes could function in part to regulate the stoichiometry of the gene products associated with its members (\cite{li_12}).  This is concordant with our results which suggest that such motifs may robustly induce hierarchical structure in the network steady-state distributions, and results from other studies suggest that such structure could be important biologically (for instance, Bayesian networks have proved to be a useful model class in modeling both gene expression data, \cite{friedman_00}, and proteomic data, \cite{sachs_05}).  Our findings also suggest another role for hierarchical coregulation motifs, namely global network stability, and provide a means of interpreting observed parameters within this context (for instance, which models will result in subcritical dynamics, and which in chaotic dynamics).

One short-coming of our analysis is the fact that we have concentrated on small-scale network motifs, and have only considered large networks which are built by combining these stochastically without further structural constraints.  Analyses of the global structure of transcription factor networks, for instance using ENCODE data to study the human TF network (\cite{gerstein_12}), has suggested that it may be appropriate to view the global structure as layered (for instance, by categorizing TFs as high-level, mid-level or low-level regulators), and that the occurrence of network motifs is highly constrained by this global topological structure.  Alternatively, scale-free global topological structure has been shown to be ubiquitous in biological networks (\cite{aldana_03}).  It is likely that incorporating such interactions between global topological structure and motif occurrence will substantially modify the analysis given here.  However, we believe that the simplified case presented here is important as a first step, and may also be of relevance from an evolutionary stand-point, since plausibly the evolution of isolated network motifs is expected to precede (and enable) the evolution of global structure (such predictions may be tested via simulations using an approach such as \cite{torres_12} for example).  A further caveat to our analyses is that they have focussed mainly on the large-network case (as $N\rightarrow\infty$) necessary to apply mean-field techniques.  As our simulations show, the properties of such motifs in small networks need not match the large-scale behaviour (for instance, we show that the multiple input module group is stabilizing in the small network context, although in the large-scale limit its mean-field dynamics match those of a network without coregulation).  Further, although we show that certain properties are robust across both Boolean and Markov Jump process model classes, further work is required to establish the generality of these results outside the Boolean context.

We hope then that this analysis provides a basis for understanding the properties of coregulation motifs in the context of global network dynamics.  As discussed, we believe that such considerations may be important in interpreting and predicting the empirically observed parameter ranges of particular motifs, and suggestive of their functionality.  Further, we hope that these results will provide a basis for future theoretical work, particularly concerning the relationship between coregulatory network motifs, global topological structure and network dynamics as discussed above, and also modeling the role of coregulatory motifs in the evolution of network structure.

\appendix

\section*{Appendix A. Proof of Proposition 1}\label{sec:appAA}

\vspace{0.3cm}
\noindent \textbf{Proof.}  We first note that the distribution $P(R')$ in the proposition is well defined, since the set of functions $R_g$ and $R_h$ for which $[\forall{k}(m(R_g(k))=R'(k))]$ is identical for any $g,h$ and fixed $R'$ (which may be denoted $\mathcal{R}_{R'}=\{R|[\forall{k}(m(R(k))=R'(k))]\}$), and any member of $\mathbb{P}(G,g,h)$ is a 1-1 mapping on this set which preserves the probability assigned to each member (by the permutational invariance property), entailing that $\sum_{R_g}P(R_g)[\forall{k}(m(R_g(k))=R'(k))] = \sum_{R_h}P(R_h)[\forall{k}(m(R_g(k))=R'(k))]$.

The proposition may be proved by induction.  The base-case at $t=0$ is given by the initial conditions.  Assuming then for induction that $c_g(v,w,t)=c_h(v,w,t)=c(v,w,t)$ at time $t$, from Eq. \ref{eq:cbn2} we have:
\begin{eqnarray}\label{eq:cbn4}
c_g(v,w,t+1) &=& \sum_{R_g,f} P(R_g)P(f)\sum_{v',w'\in\mathbb{B}^K}c_{R_g}(v',w',t)\cdot\nonumber \\
&&([f(v')=v]\wedge[f(w')=w])\nonumber\\
&=& \sum_{R_g,f} P(R_g)P(f)\sum_{v',w'\in\mathbb{B}^K}(\prod_k c_{m(R_g(k))}(v'_k,w'_k,t))\cdot\nonumber \\
&&([f(v')=v]\wedge[f(w')=w]),
\end{eqnarray}
where we use the simplicity property and the induction hypothesis to factorize $c_{R_g}(v',w',t)$.  By inspection, we have:
\begin{eqnarray}\label{eq:cbn5}
c_{m(R_g(k))}(v'_k,w'_k,t) = c_{R'(k)}(v'_k,w'_k,t),
\end{eqnarray}
for arbitrary $v',w',k$, where $R'$ is chosen such that $R_g\in\mathcal{R}_{R'}$.  Hence $\sum_{v',w'\in\mathbb{B}^K}$ $(\prod_k c_{m(R_g(k))}$ $(v'_k,w'_k,t))\cdot$ $([f(v')=v]\wedge[f(w')=w])$ is identical for all such $R_g$ and independent of $g$, as is the sum when weighted by $\sum_{R_g\in\mathcal{R}_{R'}}P(R_g)=P(R')$.  The sum across $R_g$ in Eq. \ref{eq:cbn4} may thus be replaced by a sum across $R'$, leading to the update in Eq. \ref{eq:cbn3}, which is independent of $g$.

\begin{flushright}
$\square$
\end{flushright}

\section*{Appendix B. Small network simulations}\label{sec:appA}

The analysis of stability in the models above has used a mean-field approximation to network dynamics, which holds only in the case of large networks (as $N\rightarrow\infty$).  Here, we analyse the stability of all models in the small network context through simulations.

\begin{figure*}[t]
\begin{center}
\includegraphics[width = 0.9\columnwidth]{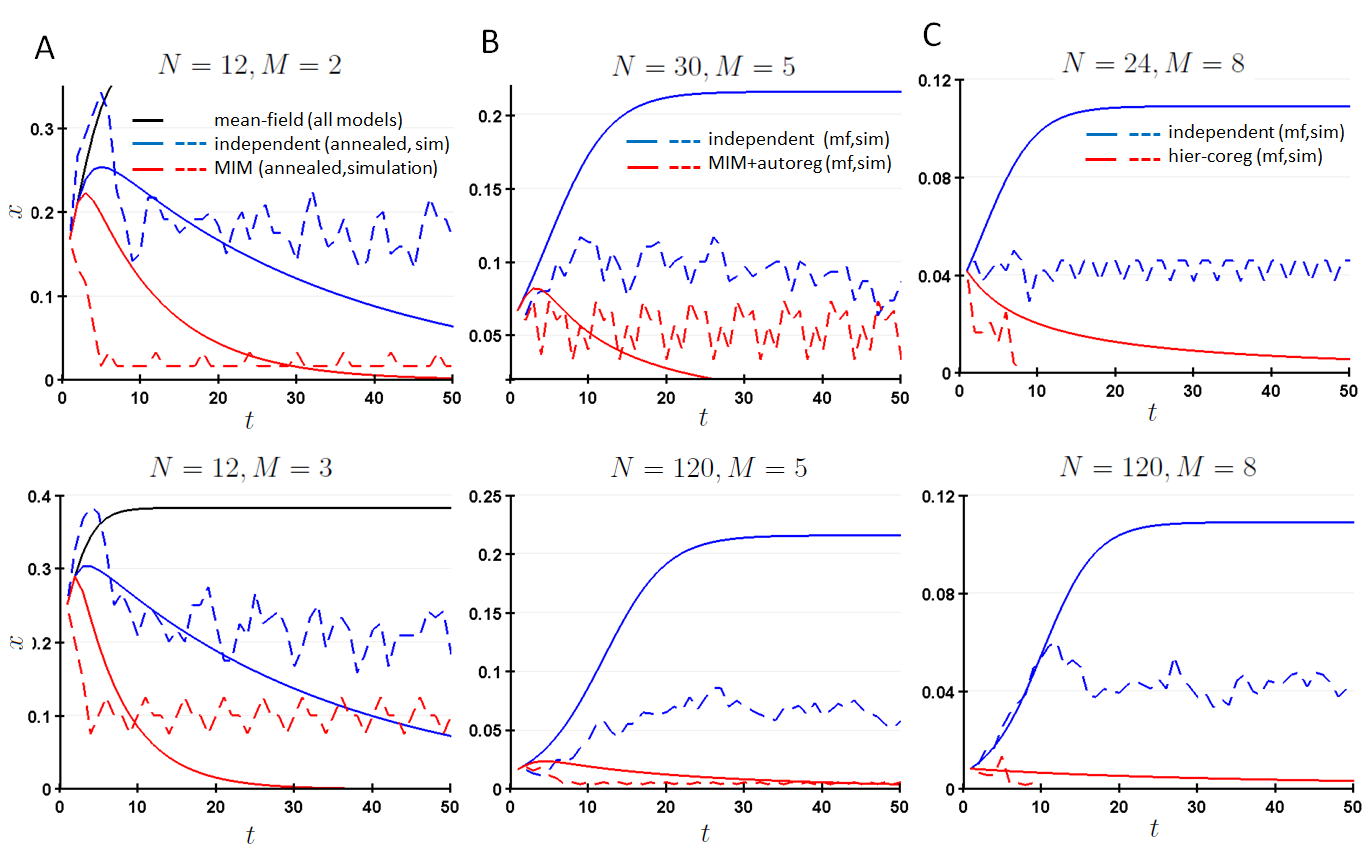}
\end{center}
\caption{{\bf Small network simulations.} {\bf (A)} Small network simulations comparing the MIM model without module groups with an independent model with matching activation frequency.  Black solid lines show mean-field approximation (same for both models).  Solid and dotted lines show expected value of $x$ (the normalized Hamming distance) calculated from the annealed approximation to each model, and the mean value of $x$ across 10 simulations respectively.  Parameter settings as shown, with $p=0.5,K=3$.  {\bf (B)} As for (A), but compares MIM model with autoregulation to matching independent model.  Solid lines show mean-field approximation.  Parameter settings as shown with $p=0.7,p_0=0.5,p_1=0.95,K=3$.  {\bf (C)} As for (B), but compares hierarchical coregulation model with matching independent model.  Parameters as shown with $p=0.5,K=6$.
}
\label{fig3}
\end{figure*}

As noted in Section \ref{sec:mim1}, the MIM model without autoregulation has identical mean-field dynamics to an $NK$-network without coregulation with matching activation frequency (which we will call the {\em independent model}, and where activation frequency denotes the expected proportion of positive rule outputs).  This does not necessarily imply the models have equal stability however for small network sizes.  To investigate this, we ran simulations of the MIM model without autoregulation, setting $N=12$, $K=3$, $p=0.5$ and $M=2,3$.  For convenience, we set $L=1$, so there is one module per module group, and $q=1$, so the activation of each group occurs with probability $p=0.5$.  We compared this model to an independent model with $N=12$, $K=3$, $p=0.5$, and do not assume the simplicity condition on $R$ for either model.  We ran 10 pairs of simulations of each model, where in each case one member of the pair is initialized to a random network state uniformly over $\mathbb{B}^N$, and the other is initialized by flipping the state of $M$ genes of the first.  We then track $x$, the Hamming distance normalized by $N$, for $T=50$ time-steps.  Figure \ref{fig3}A compares the mean value of $x$ across all simulations for $M=2$ (upper), and $M=3$ (lower) (dotted lines show simulation results, and black line shows the mean-field estimate).  As shown, the mean-field estimate severely over-estimates $x$ for $t>\sim5$.  Further, the simulations show that in this small network context, the MIM model does achieve greater stability (lower $x$, implying smaller Hamming distance) than the independent model.

The greater stability of the MIM model above in the small network context can be analysed by considering the full dynamics of the annealed model, as opposed to a mean-field approximation.  Since for $L=1$, $q=1$, all genes in a coregulated group must be either $0$ or $1$ together, we can consider such a MIM model to be equivalent to an independent model with $N/M$ nodes.   For an independent model with $N$ nodes, $x$ can take the values $0,1/N,2/N,...,1$ at each time-step.  In the annealed model (where $R$ and $f$ are resampled at each time-step), we can calculate exactly the distribution at each time step by setting:
\begin{eqnarray}\label{eq:appA1}
P(x_{t+1}) = \frac{1}{N}\sum_{x_t} P(x_t)\Binomial(Nx_{t+1};N,2p(1-p)(1-(1-x_t)^K)),
\end{eqnarray}
where $\Binomial(.;N,p)$ is the Binomial distribution with $N$ trials and mean $Np$.  Using Eq. \ref{eq:appA1}, we can derive an exact expression for the expected value of $x$ under the annealed model:
\begin{eqnarray}\label{eq:appA2}
\mathbb{E}[x_{t+1}] &=& \sum_{x_{t+1}} P(x_{t+1})x_{t+1} \nonumber\\
&=& \frac{1}{N}\sum_{x_{t+1}} x_{t+1} \sum_{x_t} P(x_t)\Binomial(Nx_{t+1};N,2p(1-p)(1-(1-x_t)^K)) \nonumber \\
&=& \sum_{x_t} P(x_t) \mathbb{E}[x_{t+1}|x_t],
\end{eqnarray}
where
\begin{eqnarray}\label{eq:appA3}
\mathbb{E}[x_{t+1}|x_t] = 2p(1-p)(1-(1-x_t)^K).
\end{eqnarray}
In Figure \ref{fig3}A, the exact expectations for $x_t$ using Eq. \ref{eq:appA2} are plotted as solid blue/red lines for the independent/MIM model respectively, which are shown to provided better approximations to the simulations than the mean-field dynamics.

Since $\mathbb{E}[x_{t+1}|x_t]$ is concave in $x_t$ (as $(1-x_t)^K$ is convex), by Jensen's inequality we have:
\begin{eqnarray}\label{eq:appA4}
\mathbb{E}[x_{t+1}] \leq \mathbb{E}\left[x_{t+1}|\mathbb{E}[x_t]\right].
\end{eqnarray}
In the following, we assume that $N>>(N/M)$, and that $P(x_1)$ is a delta distribution (we choose a fixed Hamming distance $Nx_0$ at initialization).  Then, at $t=2$, $P(x_t)=(1/N)\Binomial(.;N,c)$ for the independent model, and $P(x_t)=(M/N)\Binomial(.;N/M,c)$ for the MIM model, where $c=2p(1-p)(1-(1-x_0)^K)$, and for both models, $\mathbb{E}(x_2)=c$.  Since $N>>(N/M)$, $P(x_2)$ is highly peaked for the independent model, and hence the upper-bound in Eq. \ref{eq:appA4} is approximately satisfied, leading to $x_{3}\approx c'$, where $c'=2p(1-p)(1-(1-c)^K)$.  For the MIM model, $P(x_2)$ is less highly peaked, and hence $\mathbb{E}[x_3]<c'$.  For $t>3$, we can show by induction that the expected value of the MIM model remains below that of the independent model, since:
\begin{eqnarray}\label{eq:appA5}
\mathbb{E}_{\text{MIM}}[x_{t+1}] &\leq& \mathbb{E}_{\text{MIM}}\left[x_{t+1}|\mathbb{E}_{\text{MIM}}[x_t]\right] \nonumber\\
&<& \mathbb{E}_{\text{indep}}\left[x_{t+1}|\mathbb{E}_{\text{indep}}[x_t]\right] \nonumber\\
&\approx& \mathbb{E}_{\text{indep}}[x_{t+1}],
\end{eqnarray}
where we write $\mathbb{E}_{\text{indep}}[.]$ and $\mathbb{E}_{\text{MIM}}[.]$ for the expected values under the independent and MIM models (with $N$ and $N/M$ nodes) respectively, and use Jensen's inequality in the first line, the inductive hypothesis in the second, and the fact that $N>>M$ in the third.  Eq. \ref{eq:appA5} is confirmed in Fig. \ref{fig3}A, and provides a rationale for the greater stability of the MIM model in the small network simulations.

We further investigate the MIM model with autoregulation (see Section \ref{sec:mim2}) in the small network setting in Fig. \ref{fig3}B.  Here, we choose the parameters $p=0.7,p_0=0.5,p_1=0.95,K=3,M=5$, for which the mean-field analysis predicts greater stability for the coregulated model, which has only a zero fixed-point, than the matching independent model, which has a non-zero fixed point.  We compare networks with $N=30$ nodes (upper) and $N=120$ nodes lower (for smaller networks, the variation in network dynamics across simulations was significantly larger, and the models were not sharply distinguished).  As expected, in the larger $N=120$ network, simulations of the independent and coregulated models are sharply separated, tending towards non-zero and zero fixed points respectively.  Simulations for $N=30$ show that this behaviour also occurs in the smaller network, although the separation of the models is less pronounced.  In Fig. \ref{fig3}C we simulate the hierarchical coregulation model (see Section \ref{sec:coreg}), again choosing a parameter setting for which the mean-field analysis predicts a zero fixed-point only for the coregulated model, and a non-zero fixed point for the matching independent model ($p=0.5,K=6,M=8$).  We compare $N=24$ (upper) and $N=120$ (lower) network sizes.  In both cases, the coregulated model tends quickly to the zero fixed-point in all simulations.  The independent model simulations exhibit non-zero fixed-points, with the larger network fixed-point being higher, providing a better fit to the predicted mean-field dynamics.  As a whole, the simulations above suggest that the mean-field analysis remains informative for small networks, although the differences predicted between models may be less pronounced than for larger networks, and new phenomena may emerge, which the analysis of the exact dynamics in the annealed model sheds light on.

\section*{Appendix C. Proof of Proposition 3}\label{sec:appB}

\vspace{0.3cm}
\noindent\textbf{Proof.}  For the autoregulated MIM model with $L=q=1$, the only valid settings of the coregulated groups are $S=\{[0\;\mathbf{0}_{M-1}],[1\;\mathbf{0}_{M-1}],
[0\;\mathbf{1}_{M-1}],$ $[1\;\mathbf{1}_{M-1}]\}$, where $\mathbf{0}_A$ is the vector of $0$'s length $A$ (similarly for $\mathbf{1}_A$).  Using Proposition 1 and the above, we therefore need only consider $c(v,w,t+1)$ for $(v,w)\in S^2$ to derive the mean-field dynamics for the autoregulated model. We will write $c_1(v_1,w_1,t)$ for $P([\sigma_{C(g,1)}(1,t)=v_1]\wedge[\sigma_{C(g,1)}(2,t)=w_1])$, $c_2(v_2,w_2,t)$ for $P([\sigma_{C(g,2...M)}(1,t)=v_2]\wedge[\sigma_{C(g,2...M)}(2,t)=w_2])$, and $u(t) = c_1(0,0,t)$, $v(t)=c_1(1,1,t)$, $y(t) = c_2(0,1,t)+c_2(1,0,t)$.  Then, using Eq. \ref{eq:mim6} and writing $H(.,.)$ for the Hamming distance, we have:
\begin{eqnarray}\label{eq:appB1}
x(t) &=& \frac{1}{M}\sum_{v,w\in\mathbb{B}^m}c(v,w,t)H(v,w)\nonumber \\
&=& \frac{1}{M}((1-u(t)-v(t))+(M-1)y(t)),
\end{eqnarray}
and the updates for $u,v,y$ take the form:
\begin{eqnarray}\label{eq:appB2}
u(t+1) &=& u(t)(\neg p_0\neg x^{K-1} + \neg p_0^2(1-\neg x^{K-1}) - \neg p_1\neg p_0) + \nonumber \\
&& v(t)(\neg p_1\neg x^{K-1} + \neg p_1^2(1-\neg x^{K-1}) - \neg p_1\neg p_0) + \neg p_1\neg p_0 \nonumber \\
v(t+1) &=& u(t)(p_0\neg x^{K-1} + p_0^2(1-\neg x^{K-1}) - p_1 p_0) + \nonumber \\
&& v(t)(p_1\neg x^{K-1} + p_1^2(1-\neg x^{K-1}) - p_1 p_0) + p_1p_0 \nonumber \\
y(t+1) &=& 2p(1-p)(1-(u(t)+v(t))\neg x^{K-1}),
\end{eqnarray}
where we write $\neg p,\neg p_0,\neg p_1,\neg x$, for $(1-p),(1-p_0),(1-p_1),(1-x(t))$ respectively.  At a fixed-point, we must have $u(t+1)=u(t),v(t+1)=v(t),y(t+1)=y(t)$, which jointly imply $x(t+1)=x(t)$.  Hence, by eliminating $y$, we can show by algebraic manipulation that at steady state the following relationship must hold:
\begin{eqnarray}\label{eq:appB3}
x &=& g(x) \nonumber \\
g(x) &=& \frac{1}{M}\left(1-Z + (M-1)2p(1-p)(1-Z\neg x^{K-1})\right),
\end{eqnarray}
where
\begin{eqnarray}\label{eq:appB4}
Z &=& u + v \nonumber \\
u &=& Au + Bv + \neg p_1\neg p_0 \nonumber \\
v &=& Cu + Dv + p_1 p_0 \nonumber \\
A &=& \neg x^{K-1}\neg p_0p_0 + \neg p_0^2 - \neg p_1\neg p_0 \nonumber \\
B &=& \neg x^{K-1}\neg p_1p_1 + \neg p_1^2 - \neg p_1\neg p_0 \nonumber \\
C &=& \neg x^{K-1}\neg p_0p_0 + p_0^2 - p_1 p_0 \nonumber \\
D &=& \neg x^{K-1}\neg p_1p_1 + p_1^2 - p_1 p_0. \nonumber \\
\end{eqnarray}
Writing $g'(x)$ and $Z'(x)$ for the differentials of $g(x)$ and $Z(x)$ with respect to $x$ respectively, we can show:
\begin{eqnarray}\label{eq:appB5}
g'(x) &=& \frac{1}{M}\left(2p\neg p(M-1)(K-1)Z\neg x^{K-2} - Z'(x)\cdot(1+2p\neg p(M-1)\neg x^{K-1})\right) \nonumber \\
g'(0) &=& \frac{1}{M}\left(2p\neg p(M-1)(K-1) - Z'(0)(1+2p\neg p(M-1))\right),
\end{eqnarray}
where, noting that $Z,A,B,C,D$ are all implicitly functions of $x$,
\begin{eqnarray}\label{eq:appB5a}
Z'(x) = \frac{(p_1p_0-\neg p_1\neg p_0)(B'-A')}{(1-D)(1-A)-BC} + Z\frac{A'(1+B-D)+B'(1+C-A)}{(1-D)(1-A)-BC},
\end{eqnarray}
leading to $Z'(0)$ as in Eq. \ref{eq:mim9}, and in Eq. \ref{eq:appB5} we have used the fact that $Z(0)=1$ (for zero Hamming distance, $u+v=1$).  Hence, the MIM model with autoregulation will be stable (will not have a fixed-point for $x>0$) iff $g'(0)<1$, so long as $g(x)$ is concave increasing over the range $[0\;1]$.  By inspection of Eq. \ref{eq:appB3}, $g(x)$ is concave increasing whenever $Z$ is convex decreasing (since $(1-x)^{K-1}$ is convex decreasing and positive, and taking the product of two convex, decreasing (or increasing), positive functions preserves convexity).  This can be ascertained by calculating $Z'$ (using Eq. \ref{eq:appB5}) and $Z''$ (by differentiating Eq. \ref{eq:appB5a} again with respect to $x$) across this range for particular settings of $p,p_0,p_1,K$, to confirm that $Z'\geq 0$ and $Z''$ is decreasing.  We found this to be the case for all parameter settings used in this paper.

By substituting Eq. \ref{eq:mim7} into Eq. \ref{eq:rbn4}, we have that the independent model with matching activation frequency is stable iff $2Kp'(1-p')<1$, where $p'=(1/M)(0.5(p_0+p_1)+(M-1)p)$. Hence, under the condition that $Z(x)$ is convex decreasing over $[0\;1]$ (as stated in the proposition), the MIM model with autoregulation is at least as stable as the matching independent model when:
\begin{eqnarray}\label{eq:appB6}
g'(0) &\leq& 2Kp'(1-p'),
\end{eqnarray}
and the independent model is at least as stable as the MIM model when $g'(0) \geq 2Kp'(1-p')$.  Eq. \ref{eq:mim8} follows by substituting Eq. \ref{eq:appB5} into Eq. \ref{eq:appB6} and rearranging.

\begin{flushright}
$\square$
\end{flushright}

\section*{Appendix D. Comparing Canalyzing and Hierarchical Coregulation rules}\label{sec:appC}

In \cite{kauffman_04}, the notion of a {\em nested canalysing rule} is defined, and it is noted that regulatory interactions in many observed transcription networks can be cast in this form. A nested canalysing rule, $f:\mathbb{B}^K\rightarrow\mathbb{B}$ is such that, there exists a vector of input values, $[I_1,I_2,...,I_K]\in\mathbb{B}^K$, and output values, $[O_1,O_2,...,O_K,O_{K+1}]$, where:
\begin{eqnarray}\label{eq:appC}
f(v) &=&  \begin{cases} O_1 &\mbox{if } v(1)=I_1 \\
O_2 &\mbox{if } v(1)\neq I_1,\; v(2)=I_2 \\
&\mbox{...} \\
O_K &\mbox{if } v(1)\neq I_1,\; v(2)\neq I_2 ... v(K)=I_K \\
O_{K+1} & \mbox{otherwise,}
\end{cases}
\end{eqnarray}
Hence, if the first input takes the {\em canayzing value}, $I_1$, the output $O_1$ is generated regardless of the other inputs; if the first input does not take $I_1$, but the second takes its canalyzing value, $I_2$, the output is $O_2$ regardless; and so on, with $O_{K+1}$ being generated if no input takes its canalyzing value.

Nested canalyzing rules are similar to hierarchical coregulation rules (see Section \ref{sec:hiercoreg}) in that they place a hierarchical structure over the genes involved in the rule.  However, where canalyzing rules place a hierarchy over the rule inputs, hierarchical coregulation rules place it over the outputs.  A tighter analogy can be drawn as follows:  Consider the hierarchical coregulation model as in Section \ref{sec:hiercoreg}, with a total order over the outputs, $\Pa(m) = m-1$.  Then, a way to approximate a valid coregulation rule for group $g$ is as follows: First, select $J$ regulators for $C(g,1)$ (the first node in the group), and select a nested canalyzing rule $f_1:\mathbb{B}^J\rightarrow \mathbb{B}$ for this node.  Second, for node $C(g,2)$, select $C(g,1)$ as the first regulator and $J$ further regulators, and select a nested canalysing rule of the form $f_2:\mathbb{B}^{J+1}\rightarrow \mathbb{B}$ with $I_1=O_1=0$, and the remaining parameters arbitrary.  This has the effect that, when $C(g,1)$ is $0$, $C(g,2)$ is forced to be $0$ also.  For node $C(g,3)$, select nodes $C(g,1)$ and $C(g,2)$ as the first two regulators and add $J$ further regulators, and select a nested canalyzing rule with $I_1=O_1=I_2=O_2=0$, and the remaining parameters arbitrary, and so on for the remaining nodes $C(g,4)...C(g,K)$.

The union of the final $J$ regulators for each node in group $g$ can be considered to be the `proper regulators' of the group, which for convenience we assume do not belong to $g$, and may not all be distinct.  If we fix these proper regulators to an arbitrary setting and apply updates to group $g$, after (at most) $M$ time-steps we will reach a valid setting of the nodes in group $g$ (in the sense of the hierarchical coregulation model), such that $\sigma_{C(g,m)}=1$ only if $m=1$, or $\sigma_{C(g,m-1)}=1$.  However, if we do not fix the proper regulators as above, but allow all nodes in the network to update at each time-step, we cannot expect this property to hold at any particular time-step.  The above suggests that the hierarchical coregulation model can alternatively be viewed in terms of a separation of time-scales: the coregulatory relationships between a rule's outputs can be considered to be regulatory relationships acting at a faster time-scale (instantaneously) compared to those between the rule's inputs and outputs.  In this sense, the hierarchical coregulation model is implicitly a multi-scale model.

\section*{Appendix E. Markov Jump Process Simulations for Hierarchical Coregulation Model}\label{sec:appD}

As mentioned in Section \ref{sec:mjp}, in the Markov Jump process analogues of the independent and hierarchical coregulation models, we let $\mathbf{x}\in(\mathbb{N}\cup\{0\})^N$, which can be taken to represent for instance the transcript count associated with each gene at a given point in time.  We write $\mathbf{x}(t)$ for the joint setting of transcript counts at time $t$, where $t\in[0\;T]$, and hence varies continuously.  As noted, we retain all parameters from the Boolean models, including $N,K,M,p,P(R),P(f)$ for the hierarchical coregulation model. In addition, we introduce the parameters $a,b,d$, where $a$ and $b$ are low and high expected steady-state transcript counts which are used to represent a given gene being on or off respectively, and $d$ is a degradation rate common to all genes.  We assume that $a\approx 0$, and that $b>>a$.  For our simulations, we take $a=0.1$, $b=20$, $d=0.01$.

Each rule in the Boolean model becomes a set of rate equations in the MJP model.  In all simulations, we use $K=2$, hence each rule has 2 inputs.  Additionally, in the coregulation model, we split each group rule over $M$ outputs into $M$ rules with a single output each (hence, while the group responses are tied by $P(f)$, we do not model joint stochasticity across the group).  For a given gene $Y$ then, with regulators $X_1$ and $X_2$ and update rule $f_Y:\mathbb{B}^2\rightarrow\mathbb{B}$ derived from the Boolean model, we introduce the following rate equations:
\begin{alignat}{3}\label{eq:chemEq}
\emptyset &\xrightarrow{k_{00}} Y &\quad
X_1 &\xrightarrow{k_{10}} X_1 + Y &\quad
X_1+Y &\xrightarrow{k_{10}'} X_1  \nonumber \\
Y &\xrightarrow{k_{00}'} \emptyset &\quad
X_2 &\xrightarrow{k_{01}} X_2 + Y &\quad
X_2+Y &\xrightarrow{k_{01}'} X_2  \nonumber \\
X_1+X_2 &\xrightarrow{k_{11}} X_1+X_2+Y &\quad
& &\quad
X_1+X_2+Y &\xrightarrow{k_{11}'} X_1+X_2
\end{alignat}
where
\begin{eqnarray}\label{eq:appD1}
k_{00} &=& ad[f_{00}=0] + bd[f_{00}=1] \nonumber \\
k_{10} &=& (1/b)(b-a)d[f_{10}=1\wedge f_{00}=0] \nonumber \\
k_{10}' &=& d((1/a)-(1/b))[f_{10}=0\wedge f_{00}=1] \nonumber \\
k_{01} &=& (1/b)(b-a)d[f_{01}=1\wedge f_{00}=0] \nonumber \\
k_{01}' &=& d((1/a)-(1/b))[f_{01}=0\wedge f_{00}=1] \nonumber \\
k_{11} &=& (1/b^2)(br'-r)[f_{11}=1\wedge (r'/r)<b] \nonumber \\
k_{11}' &=& (1/b^2)(((r/b)-r')[f_{11}=1\wedge (r'/r)>b] + ((r/a)-r')[f_{11}=0]),\nonumber\\
\end{eqnarray}
and $f_{xy}=f([x\;y])$, $r=k_{00}+bk_{10}+bk_{01}$ and $r'=d+bk_{10}'+bk_{01}'$.

Following \cite{wilkinson_11} and \cite{gillespie_07}, each reaction in Eq. \ref{eq:chemEq} can be associated with a propensity function $\alpha$ whose value is the product of the reactant counts at a given time and the reaction rate constant; hence, numbering the reactions across and then down, $\alpha_1(t) = k_{00}$, $\alpha_2(t)=X_1(t)k_{10}$, $\alpha_3(t)=X_1(t)Y(t)k_{10}'$, and so on.  The probability that reaction $j$ occurs within a small time interval $[t,t+\text{d}t)$ is $\alpha_j(t)\text{d}t$, and the model is efficiently simulated using the Gillespie algorithm (\cite{gillespie_07}).

The parameter settings above are chosen such that if the counts of the regulators $X_1,X_2$ are fixed at the levels $(0,0),(b,0),(0,b),(b,b)$, the expected count of $Y$ at steady-state will be $f^*(0,0),f^*(1,0),f^*(0,1),f^*(1,1)$ respectively, where $f^*(x,y)=a$ if $f(x,y)=0$, and $f^*(x,y)=b$ if $f(x,y)=1$.  We illustrate by evaluating the expression levels of $Y$ for the XOR function, $f(x,y)=(x\vee y)-(x\wedge y)$.  For $X_1(t)=0,X_2(t)=0$, only reactions 1 and 4 occur, hence $Y$ is produced at rate $ad$ and degraded at rate $Y(t)d$, leading to an expected count at steady-state of $ad/d=a$.  For $X_1(t)=b,X_2(t)=0$, reactions 1,2 and 4 occur (3 does not, since the condition $[f_{10}=0\wedge f_{00}=1]$ is not satisfied in Eq. \ref{eq:appD1}, and so $k_{10}'=0$).  Reaction 2 occurs at rate $X_1(t)(1/b)(b-a)d=b(1/b)(b-a)d=bd-ad$, and so the combined rate at which $Y$ is produced from reactions 1 and 2 is $bd-ad+ad=bd$, and the steady-state expected count is $bd/d=b$.  For $X_1(t)=0,X_2(t)=b$, reactions 1, 4 and 5 occur, and a similar analysis to $X_1(t)=b,X_2(t)=0$ leads to an expected count of $b$ for $Y$.  For $X_1(t)=b,X_2(t)=b$, reactions 1, 2, 4, 5 and 8 occur, with the rates of 1, 2, 4 and 5 as above, and reaction 8 occurs at a rate $X_1(t)X_2(t)Y(t)(1/b^2)((r/a)-r')$.  The rate of production of $Y$ is therefore $2bd-2ad+ad=2bd-ad$, and writing $\delta(t)$ for the rate of degradation of $Y$ at time $t$, we have:
\begin{eqnarray}
\delta(t) &=& Y(t)(d + X_1(t)X_2(t)(1/b^2)((r/a)-r')) \nonumber\\
&=& Y(t)(d + (b^2)(1/b^2)(((ad+2bd(b-a)/b)/a)-d) \nonumber\\
&=& Y(t)(ad+2d(b-a))/a,
\end{eqnarray}
so that the steady-state expectation is $(2bd-ad)a/(ad+2d(b-a)) = a = f^*(1,1)$.  The chosen rate constants can similarly be shown to mimic all possible Boolean function settings of $f$ by calculations similar to those above.  For $a\approx 0$, the calculations above will hold approximately when the regulators are expressed at levels $(a,a),(b,a),(a,b),(b,b)$, allowing a network which combines such rules to approximate the dynamics of the associated Boolean model.



%
%
%

\end{document}